\documentclass[lettersize,journal]{IEEEtran}
\usepackage{array}
\usepackage{multicol}
\usepackage[caption=false,font=normalsize,labelfont=sf,textfont=sf]{subfig}
\usepackage{textcomp}
\usepackage{stfloats}
\usepackage{url}
\usepackage{verbatim}
\usepackage{graphicx}
\usepackage{tikz}
\usepackage{graphicx}
\usepackage{cite}
\usepackage{multirow}
\usepackage{makecell}
\usepackage[ruled,linesnumbered]{algorithm2e}
\usepackage{amsmath}
\usepackage{bm}
\usepackage{color}
\usepackage{graphicx}
\usepackage{amssymb}
\usepackage{url}
\usepackage{booktabs}
\usepackage{multirow}
\usepackage[normalem]{ulem}
\usepackage{ragged2e}
\usepackage{soul}
\usepackage{xcolor}
\usepackage{ulem}
\usepackage{colortbl}
\usepackage{booktabs}
\usepackage[normalem]{ulem}
\useunder{\uline}{\ul}{}
\usepackage[colorlinks,linkcolor=blue,citecolor=blue]{hyperref}
\usepackage{longtable}
\usepackage{relsize}
\usepackage{comment}
\usepackage{diagbox}
\newcolumntype{C}[1]{>{\centering\arraybackslash}p{#1}}
\newcolumntype{L}[1]{>{\raggedright\arraybackslash}p{#1}}

\begin{document}
\title{
Contrastive Representation-Guided Genetic Minority Oversampling for Imbalanced Time-Series Classification}



\author{
Wenbin~Pei,
Yunrong~Hao,
Zhen~Liu,
Guan~Wang,
Bing~Xue,~\IEEEmembership{Fellow,~IEEE},
Yiu-Ming~Cheung,~\IEEEmembership{Fellow,~IEEE},
and Qiang~Zhang,~\IEEEmembership{Senior~Member,~IEEE}
\thanks{This work was supported in part by the National Key Research and Development Program of China under grant 2024YFA1012700, the National Natural Science Foundation of China under grants 62206041, 12371516 and U21A20491, and the NSFC-Liaoning Province United Foundation under grant U1908214, the 111 Project under grant D23006, the Liaoning Revitalization Talents Program under grant XLYC2008017, and China University Industry-University-Research Innovation Fund under grants 2022IT174, Natural Science Foundation of Liaoning Province under grant 2023-BSBA-030, and an Open Fund of National Engineering Laboratory for Big Data System Computing Technology under grant SZU-BDSC-OF2024-09. (Corresponding author: Zhen Liu.)
}
\thanks{Wenbin Pei and Yunrong Hao are with the School of Computer Science and Technology, Dalian University of Technology, Dalian 116024, China; Key Laboratory of Social Computing and Cognitive Intelligence (Dalian University of Technology), Ministry of Education, Dalian 116024, China (e-mail: peiwenbin@dlut.edu.cn; haoyunrong@mail.dlut.edu.cn).}%
\thanks{Zhen Liu is with the College of Computing and Data Science, Nanyang Technological University, Singapore (e-mail: zhen.liu@ntu.edu.sg).}%
\thanks{Guan Wang is with the School of Mechanics and Aerospace Engineering, Dalian University of Technology, Dalian 116024, China (e-mail: guanwang@dlut.edu.cn).}%
\thanks{Bing Xue is with the School of Engineering and Computer Science, Victoria University of Wellington, PO Box 600, Wellington 6140, New Zealand (e-mails: bing.xue@ecs.vuw.ac.nz).}%
\thanks{Yiu-Ming Cheung is with the Department of Computer Science, Hong Kong Baptist University, Hong Kong, SAR, China (e-mail: ymc@comp.hkbu.edu.hk).}%
\thanks{Qiang Zhang is with the School of Computer Science and Technology, 
Dalian University of Technology, Dalian 116024, China; 
Key Laboratory of Social Computing and Cognitive Intelligence 
(Dalian University of Technology), Ministry of Education, Dalian 116024, China; 
and also with the National and Local Joint Engineering Laboratory of Computer Aided Design, 
Dalian University, Dalian 116622, China 
(e-mail: zhangq@dlut.edu.cn).}}%
\maketitle

\begin{abstract}

Real-world time-series classification tasks often exhibit class imbalance, which can be extremely severe in some applications. To avoid training biased classifiers on imbalanced data, sampling is one of the most popular data pre-processing techniques because of its classifier-agnostic nature. However, due to the complex temporal dependencies in original time-series data and the scarcity of minority-class samples, existing sampling methods, including interpolation-based oversampling methods and deep learning-based generative models, usually suffer from limited generalization and poor diversity when generating new time-series samples. This paper proposes a Frequency-domain representation-guided Multi-tree Genetic Programming-based oversampling approach (FreMGP) to imbalanced time-series classification, where each individual represents a set of synthetic samples for the minority class.
A frequency-domain class-discriminative representation module based on contrastive learning is also developed, guiding the evolutionary search toward high-quality synthetic time-series samples. Experiments on imbalanced time-series datasets demonstrate that FreMGP outperforms existing oversampling methods and consistently improves the performance of different classifiers, including both general machine learning and deep learning models.

\end{abstract}


\section{Introduction}



Many real-world applications generate substantial amounts of time-series data. To make sense of such data, time-series classification (TSC) has emerged as an important machine learning task that aims to assign predefined labels to sequential data based on its temporal patterns~\cite{foumani2024deep}.
To date, numerous classification algorithms, e.g., Hydra~\cite{dempster2023hydra} and spectral-aware reservoir computing (SARC)~\cite{pmlr-v267-liu25by}, have been proposed for TSC, while most of them implicitly assume a balanced data distribution across different classes. However, in many applications, such as fault detection~\cite{wang2024class}, medical diagnosis~\cite{kwak2024classification}, and economics~\cite{LI2025104025}, time-series data is naturally imbalanced, posing a challenge to conventional TSC algorithms. Note that data imbalance can bias classifiers toward the majority class and weaken their ability to recognize minority-class samples that usually are of particular interest~\cite{wang2025mitigating}.

Data sampling~\cite{chen2024survey} is one of the most popular techniques to address the class imbalance issue due to its classifier-agnostic nature. To rebalance data distribution, traditional interpolation-based oversampling methods, such as the synthetic minority over-sampling technique (SMOTE)~\cite{chawla2002smote} and its variants~\cite{fernandez2018smote,chen2024survey}, generate a number of synthetic samples by interpolating between minority-class samples and their neighbors. However, as these sampling methods were developed for tabular imbalanced data rather than time-series data, they are unable to capture the complex temporal patterns inherent in time series~\cite{zhao2022tsmote}. In the literature, to handle imbalanced time-series data, the temporal-oriented synthetic minority oversampling technique (T-SMOTE)~\cite{zhao2022tsmote} incorporates temporal characteristics during sample generation. Nevertheless, it still depends on local interpolation, which may limit the ability to preserve discriminative local variations~\cite{iglesias2023data} and frequency-domain characteristics in time series~\cite{wen2021time}. Furthermore, deep learning-based generative models, such as generative adversarial networks (GANs)~\cite{jeon2022gtgan}, variational autoencoders (VAEs)~\cite{desai2021timevae}, and diffusion-based models~\cite{yuan2024diffusionts}, have demonstrated their effectiveness in generating time-series data. Despite their strong data modeling capabilities, these methods typically demand a large number of training samples to accurately learn the complex temporal dependencies in time-series data~\cite{gonen2025datascarcity}. Unfortunately, from a practical perspective, this is not always workable because time-series samples in the minority class are usually scarce, resulting in limited generalization and poor diversity when generating new data~\cite{gonen2025datascarcity,shirvan2025deep}.

Genetic programming (GP) offers a solution for generating data~\cite{pei-dg-smote,pei2026evotfs}. Unlike black-box generative models, GP individuals are tree-structured, combining function and terminal nodes to form diverse mathematical expressions, each of which can be regarded as a time-series sample. More importantly, GP automatically learns complex transformations from the learned time-series data, allowing it to produce synthetic time-series samples that preserve the intricate temporal dependencies essential for time-series classification. However, it remains challenging for GP to generate high-quality, diverse time-series data when only a limited number of minority-class samples are available in an imbalanced time-series classification (ITSC) task. Therefore, it is necessary to investigate the potential of GP for oversampling time series in ITSC. In this paper, we propose a \textbf{Fre}quency-domain representation-guided \textbf{M}ulti-tree \textbf{G}enetic \textbf{P}rogramming-based oversampling approach (\textbf{FreMGP}), which learns a frequency-domain class-discriminative representation space and uses it to guide the evolutionary oversampling process.
Moreover, FreMGP adopts a multi-tree GP representation, where each individual generates a set of synthetic samples for one minority class rather than a single sample.

The main contributions of this paper are summarized as follows:

\begin{itemize}
    \item We design a frequency-domain contrastive representation learning module. It uses prototype-guided contrastive learning to construct a class-discriminative representation space. This provides a reliable representation space for evaluating the quality of generated synthetic samples.

    \item We design a novel multi-tree GP structure, for which both the terminal set and the function set are specifically designed, to represent frequency-domain time-series samples for oversampling. This enables a set of synthetic minority-class samples to be generated within a single GP individual, making it easy to evaluate the overall diversity of the generated sample set.  

    \item We develop a representation-guided two-stage fitness evaluation method and corresponding genetic operators. The fitness function evaluates generated samples in the learned representation space and considers both the similarity to existing minority-class samples and the diversity of the generated sample set. This fitness function guides the evolutionary search toward samples that are close to the minority-class region in the learned representation space while ensuring sufficient diversity of the generated samples.
\end{itemize}

\section{Background}
\subsection{Frequency-domain Contrastive Representation Learning}

Representation learning aims to transform raw data into an embedding space, where samples can be represented by compact and informative feature vectors~\cite{bengio2013representation}. 
Given $L$ as the sequence length, for a time-series sample $\mathbf{x}_i \in \mathbb{R}^{L}$,
a representation encoder $f_{\theta}(\cdot)$ maps it to a feature vector:
\begin{equation}
    \mathbf{h}_i = f_{\theta}(\mathbf{x}_i),
\end{equation}
where $\mathbf{h}_i$ is the learned representation of $\mathbf{x}_i$. For classification tasks, a desirable representation space is expected to preserve class-related information, so that samples from the same class are close to each other while samples from different classes are clearly separated.

\begin{figure}
\centering
\includegraphics[width=1\linewidth]{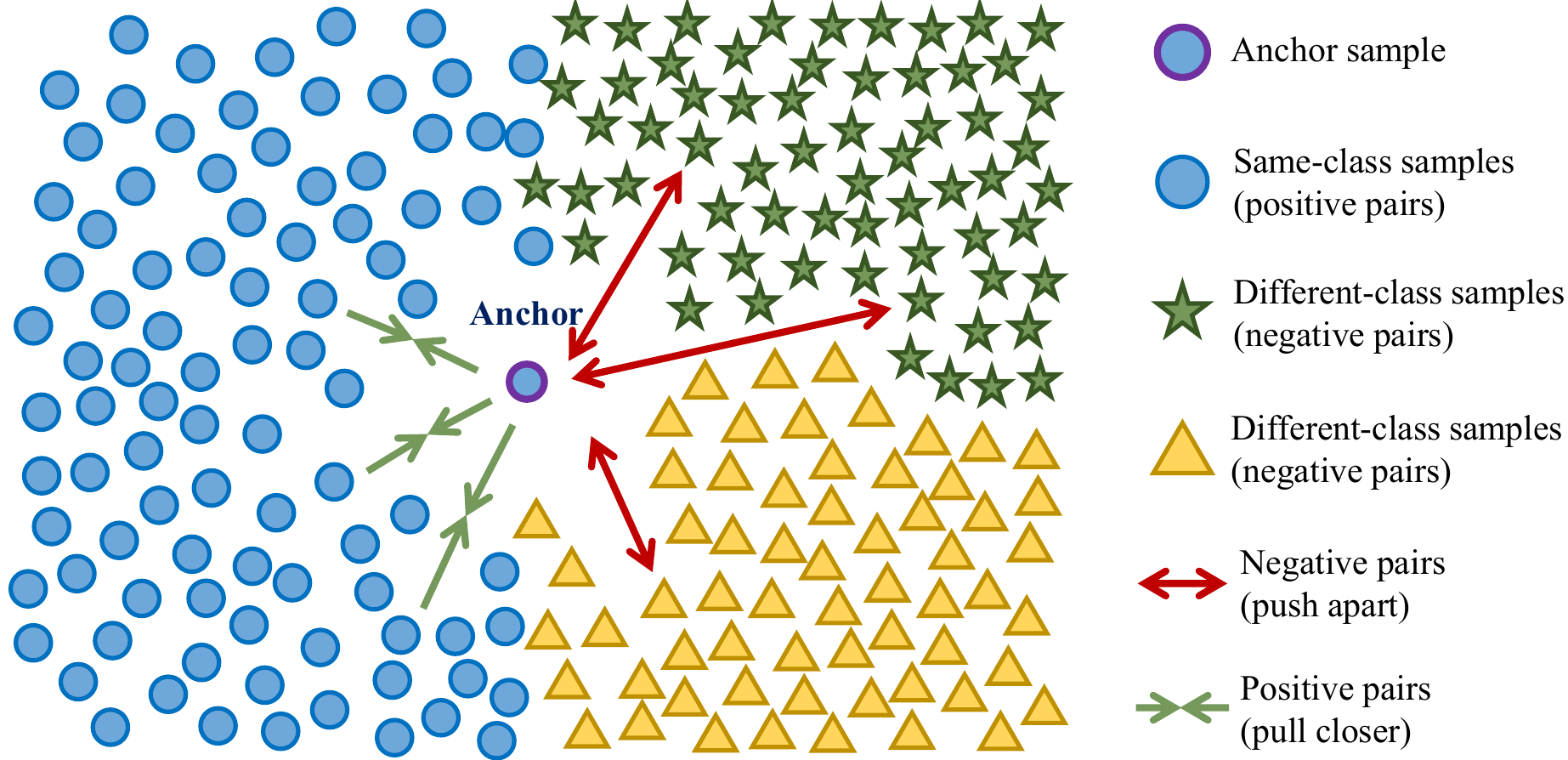}
\caption{An illustration of supervised contrastive learning in the embedding space. Samples from the same class as the anchor form positive pairs and are pulled closer, while samples from different classes form negative pairs and are pushed apart.}
\label{fig:supcon}
\end{figure}

Contrastive learning~\cite{chen2020simple,he2020momentum} is a common representation learning paradigm that learns representations by comparing positive and negative sample pairs. Positive pairs are encouraged to be close in the embedding space, while negative pairs are pushed apart. When class labels are available, supervised contrastive learning can use label information to construct positive and negative pairs~\cite{khosla2020supervised}. As illustrated in Fig.~\ref{fig:supcon}, samples from the same class as the anchor are treated as positive pairs, while samples from different classes are treated as negative pairs. 
A common supervised contrastive learning objective is defined as~\cite{khosla2020supervised}:
\begin{equation}
    \mathcal{L}_{\mathrm{com}} =
    \sum_{i=1}^{B}
    \frac{-1}{|\mathcal{P}(i)|}
    \sum_{p \in \mathcal{P}(i)}
    \log
    \frac{
    \exp(\mathbf{z}_i^\top \mathbf{z}_p / \tau)
    }{
    \sum_{j=1,\, j\ne i}^{B}
    \exp(\mathbf{z}_i^\top \mathbf{z}_j / \tau)
    },
\end{equation}
where $B$ denotes the batch size, $i$ denotes the index of an anchor sample, $\mathcal{P}(i)$ denotes the set of positive sample indices that share the same class label as $\mathbf{x}_i$ while excluding the anchor itself, $\mathbf{z}_i$ is the normalized projection of the representation $\mathbf{h}_i$, and $\tau$ is a temperature parameter.


For time-series data, representations can be learned from either the time domain or the frequency domain~\cite{Dong2024TimeSiam, woo2022cost}. Compared with raw time-domain data, frequency-domain features provide a complementary view and help capture global patterns of time series. In particular, magnitude spectra are less sensitive to temporal shifts~\cite{oppenheim2010discrete}, which is useful for learning stable representations. 
Therefore, in this work, contrastive representation learning in the frequency domain is used to construct a representation space. The learned space provides compact feature representations for time-series samples and is later used to evaluate generated samples during evolutionary oversampling.

\subsection{Multi-Tree and Strongly-Typed Genetic Programming}
Inspired by biological evolution, GP is an evolutionary algorithm that automatically evolves executable computer programs, typically represented as tree structures. In standard GP, an individual contains only one tree.  

\paragraph{Multi-Tree Genetic Programming (MTGP)}
MTGP extends the standard GP by allowing multiple trees within an individual~\cite{lensen2018generating}. An individual in MTGP can be represented as $\mathcal{T} = \{T_1, T_2, \ldots, T_M\}$, where $T_m$ denotes the $m$-th tree in the individual, and $M$ is the number of trees in this individual. 

\paragraph{Strongly-typed Genetic Programming (STGP)}
Different from standard GP, STGP enforces data type constraints on functions and terminals to construct strongly-typed trees~\cite{montana1995strongly}. In STGP, each terminal is assigned predefined data types, and correspondingly, each function has specified types for its arguments and the returned value. This restriction eliminates invalid trees, thereby reducing the search space.




In this work, MTGP is employed to enable each individual to represent a set of synthetic frequency-domain time-series samples, each of which is represented by a strongly-typed tree adhering to predefined type constraints.



\subsection{Related Works}
\paragraph{Imbalanced Time-Series Classification}
ITSC refers to the task of classifying time-series data into categories where the number of samples across different classes is imbalanced. Commonly used traditional classifiers mainly include multilayer perceptron (MLP) \cite{del2021auto}, Random Forest \cite{goehry2021random}, and distance-based methods such as k-nearest neighbors with dynamic time warping \cite{tran2019novel}. To date, deep learning has significantly advanced the field \cite{ismail2019deep}, with numerous deep models proposed to automatically learn temporal dependencies, including long short-term memory (LSTM) networks \cite{markovic2025parkinsons}, temporal convolutional networks (TCNs) \cite{zhang2023cost}, and attention-based Transformers \cite{katrompas2022recurrence}.

To address the class imbalance issue in ITSC, cost-sensitive learning has been used to improve performance on the minority class by assigning higher costs to the misclassification of minority-class instances~\cite{elkan2001foundations, zhang2023cost,wang2024class}. 
However, these cost-sensitive methods require cost matrices that are often pre-designed by humans.
Sampling\mbox{\cite{chen2024survey}} is another popular family of techniques to address the class imbalance issue due to its classifier-agnostic nature.
Sampling methods mainly include undersampling and oversampling~\cite{zhao2022tsmote, gonen2025datascarcity, pei2026evotfs}. Compared with undersampling, oversampling rebalances the class distribution by generating additional minority-class samples while retaining the original training data.

\paragraph{Existing Oversampling Methods}
The simplest oversampling is random oversampling (ROS)~\cite{batista2004study}, which simply duplicates some existing minority-class samples at random. Although ROS is simple to implement, it does not introduce new information and frequently results in overfitting. To introduce diversity, SMOTE~\cite{chawla2002smote} generates synthetic minority-class samples through linear interpolation between neighboring minority-class instances. 
Several SMOTE variants have been developed with different sampling and generation strategies, including Borderline-SMOTE~\cite{han2005borderline}, adaptive synthetic sampling approach (ADASYN)~\cite{he2008adasyn}, and grouping-based SMOTE (GB-SMOTE)~\cite{REN2023108992}. However, these methods were primarily designed for tabular imbalanced data and do not explicitly account for the temporal characteristics of time series~\cite{zhao2022tsmote}. For time-series data, T-SMOTE~\cite{zhao2022tsmote} incorporates temporal information into the oversampling process. To oversample imbalanced time-series data in ITSC, T-SMOTE\mbox{\cite{zhao2022tsmote}} extends SMOTE by incorporating temporal characteristics during sample generation. However, T-SMOTE still depends on linear interpolation, which may limit the ability to preserve discriminative local variations~\cite{iglesias2023data} and frequency-domain characteristics in time series~\cite{wen2021time}.


In addition to interpolation-based oversampling, deep learning-based generative methods have also been explored for time-series data augmentation~\cite{wen2021time,iglesias2023data}. These methods aim to learn the underlying data distribution and synthesize new samples from the learned generative model. The boundary-focused generative adversarial network (BFGAN)~\cite{lee2022boundary} introduces additional labels to reflect the importance of sample positions in the data space, so that the generator can better consider boundary information and multi-modal structures. The counterfactual augmentation minority generation (CFAMG)~\cite{wang2025mitigating} generates counterfactual minority-class samples by identifying causal factors related to class differences and intervening on latent representations. ImagenFew~\cite{gonen2025datascarcity} further studies how time series can be generated under data-scarce conditions, using a pretrained diffusion framework with dynamic convolutions and dataset-token conditioning for domain-aware generation. Although deep generative models provide a powerful approach to capturing complex time-series distribution, their use for ITSC still faces two major challenges~\cite{shirvan2025deep,gonen2025datascarcity}. First, these deep models usually require sufficient training samples to learn reliable generative distributions. However, minority-class samples are often scarce in ITSC~\cite{shirvan2025deep,gonen2025datascarcity}. Second, the generation process of deep models is usually implicit, making it difficult to precisely regulate structural properties of generated time series, such as local variations, temporal patterns, and spectral components~\cite{wen2021time,iglesias2023data}. 


The evolutionary time-frequency domain-based synthetic minority oversampling approach (Evo-TFS)~\cite{pei2026evotfs} is an evolutionary oversampling method for ITSC. It uses GP to synthesize time series from the minority class, and the evolutionary oversampling process is guided by the fitness function incorporating both time-domain and frequency-domain distances~\cite{pei2026evotfs}. Evo-TFS demonstrates the potential of GP to generate high-quality time-series samples in ITSC. However, Evo-TFS's fitness function is based on direct time-frequency similarity and finds it hard to fully capture class-discriminative information that is important for classification. Moreover, Evo-TFS requires multiple independent runs to produce the required number of minority-class samples. This makes it difficult to guarantee the overall diversity of the generated samples explicitly.
Therefore, in this work, we further investigate how GP can be used to synthesize a set of minority-class time-series samples and how these samples can be evaluated.

\section{The Proposed Approach}

\subsection{The Overall Design of FreMGP}

\begin{figure*}
\centering
  \includegraphics[width=1.0\textwidth]{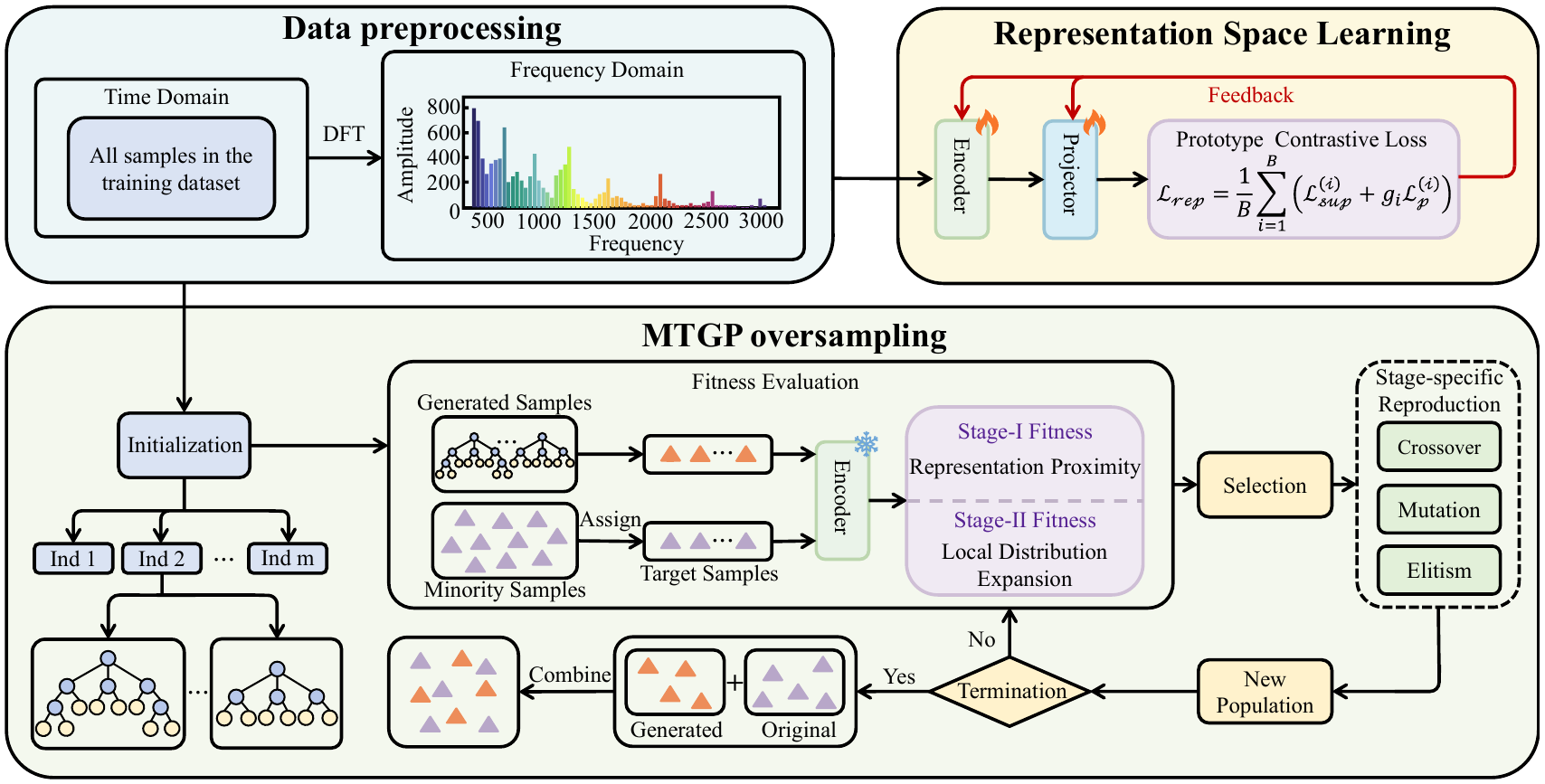}
  \caption{The Overall Design of FreMGP.
  }

  \label{fig:framework}
\end{figure*}
The goal of FreMGP is to generate high-quality and diverse minority-class time-series samples to rebalance the training data for ITSC. 
The overall design of FreMGP is described in  Fig.~\ref{fig:framework}. 

In general, representations of time-series data can be learned from either the time domain or the frequency domain \cite{Dong2024TimeSiam,woo2022cost}. Compared with time-domain data, frequency-domain features provide a complementary view and are particularly effective at capturing the global characteristics of time-series data. Moreover, compared to the time-series domain, magnitude spectra are less sensitive to temporal shifts~\cite{oppenheim2010discrete}, which benefits the learning of stable time-series representations. Therefore, the original training samples are first transformed into the frequency domain by the discrete Fourier transform (DFT)~\cite{cooley1965algorithm}, as illustrated in Fig.~\ref{fig:framework}. 
Let the training set be denoted as $\mathcal{D}_{\mathrm{train}} =
\{(\mathbf{x}_i, y_i)\}_{i=1}^{N}$, where $\mathbf{x}_i \in \mathbb{R}^{L}$ is a univariate time-series sample, $L$ is the sequence length, and $y_i$ is the class label. Each time-series sample is transformed from the time domain into the frequency domain using DFT:
\begin{equation}
    X_{i,f}
    =
    \sum_{\ell=0}^{L-1}
    x_{i,\ell}
    \exp\left(-\varsigma\frac{2\pi f\ell}{L}\right),
    \quad
    f=0,1,\ldots,F-1,
\label{eq:dft}
\end{equation}
where $X_{i,f}$ denotes the complex frequency coefficient at frequency index $f$, $x_{i,\ell}$ denotes the corresponding time-domain value at time index $\ell$, $\varsigma$ is the imaginary unit satisfying $\varsigma^2=-1$, and $F$ is the number of retained frequency components. For real-valued time series, the one-sided spectrum can be used with $F=\lfloor L/2 \rfloor+1$. The transformed frequency-domain sample is denoted as $\mathbf{X}_i \in \mathbb{C}^{F}$. Based on the frequency-domain data, FreMGP consists of two main modules:
\subsubsection{\textbf{Frequency-domain Representation Space Learning}}

In the first module, an encoder is trained through frequency-domain contrastive representation learning. This module aims to construct a class-discriminative representation space, where samples from the same class are encouraged to be close while samples from different classes are encouraged to be separated. After training, the encoder is fixed and later used in the fitness evaluation of the subsequent oversampling process.

\subsubsection{\textbf{MTGP-based Oversampling}}

In the second module, FreMGP performs evolutionary oversampling based on MTGP, where each individual contains multiple trees to represent a set of synthetic frequency-domain samples. The trained encoder is used to evaluate every individual in the learned representation space through a two-stage evolutionary search. In the first stage, generated samples are evaluated based on their similarity to their corresponding target minority-class samples. In the second stage, both the similarity and diversity are considered. The evolutionary learning process stops when the stopping criterion is satisfied.



After the evolutionary oversampling process, each generated frequency-domain sample $\widehat{\mathbf{X}}\in\mathbb{C}^{F}$ is transformed back into the time domain by the inverse discrete Fourier transform (IDFT) as $\widehat{\mathbf{x}}=\mathrm{IDFT}(\widehat{\mathbf{X}})$, where $\widehat{\mathbf{x}}\in\mathbb{R}^{L}$ is the corresponding synthetic time-series sample. Afterwards, the synthetic samples are combined with the original training set to form a rebalanced training set for downstream classifiers.
The details of the two modules in FreMGP are introduced below.
\subsection{Frequency-Domain Representation Space Learning}
Directly evaluating generated samples in the original data space may mainly capture similarity at the instance level, often failing to reflect class-discriminative information. 
Therefore, in the first module of FreMGP, we use supervised contrastive learning to construct a class-discriminative frequency-domain representation space. The learned space is then used to evaluate generated samples during the subsequent evolutionary search.
\subsection*{1) Supervised Contrastive Loss}

Specifically, for a frequency-domain sample $\mathbf{X}_i \in \mathbb{C}^{F}$, the encoder $E_{\theta}(\cdot)$ maps it into a hidden representation:
\begin{equation}
    \mathbf{h}_i = E_{\theta}(\mathbf{X}_i),
\end{equation}
where $\mathbf{h}_i \in \mathbb{R}^{d_h}$ is the learned hidden representation of $\mathbf{X}_i$.

To learn the representation space, FreMGP follows the common encoder-projector design in contrastive learning~\cite{chen2020simple,khosla2020supervised}, where a projector $P_{\phi}(\cdot)$ maps $\mathbf{h}_i$ into a normalized metric embedding:

\begin{equation}
    \mathbf{z}_i =
    \frac{P_{\phi}(\mathbf{h}_i)}
    {\|P_{\phi}(\mathbf{h}_i)\|_2},
\end{equation}
where $\mathbf{z}_i \in \mathbb{R}^{d_z}$ and $\|\mathbf{z}_i\|_2 = 1$.

Based on the normalized metric embeddings, FreMGP adopts a supervised contrastive loss\mbox{\cite{khosla2020supervised}} to learn a class-discriminative embedding space. For each anchor sample $\mathbf{X}_i$, samples with the same class label in a batch are treated as positive samples. The supervised contrastive loss is defined as follows:
\begin{equation}
    \mathcal{L}_{sup}^{(i)}
    =
    \frac{-1}{|\mathcal{P}(i)|}
    \sum_{p\in\mathcal{P}(i)}
    \log
    \frac{
    \exp(\mathbf{z}_i^{T}\mathbf{z}_{p}/\tau)
    }{
    \sum_{j\ne i}
    \exp(\mathbf{z}_i^{T}\mathbf{z}_{j}/\tau)
    },
\end{equation}
where $\mathcal{P}(i)=\{p \mid y_p=y_i, p\ne i\}$ is the positive index set of anchor sample $\mathbf{X}_i$; $\mathbf{z}_i$, $\mathbf{z}_p$, and $\mathbf{z}_j$ denote the normalized metric embeddings of the anchor sample, a positive sample, and a non-anchor sample in the batch, respectively; and $\tau$ is the temperature parameter.

\subsection*{2) Prototype-Guided Loss}

Although supervised contrastive learning encourages intra-class compactness and inter-class separation, class imbalance may still bias the learned representation space toward majority classes. This is because minority-class samples typically provide fewer same-class positives within each batch. To alleviate this issue, FreMGP introduces fixed class prototypes as class-level anchors, providing each class with a stable class-level reference beyond sample-to-sample comparisons.


The fixed prototypes are constructed in three steps. First, an initial metric embedding space is trained using the normalized temperature-scaled cross-entropy (NT-Xent) loss~\cite{chen2020simple}, from which the normalized embeddings of majority-class samples are extracted. Second, a majority-class reference prototype is optimized on the unit hypersphere by minimizing its average Euclidean distance to these embeddings. Finally, FreMGP constructs a set of simplex equiangular tight frame (ETF) prototypes, where all class prototypes are uniformly separated with equal angular distances\mbox{\cite{yang2023neural}}. The majority-class simplex prototype is then aligned with the optimized majority-class reference prototype, and the same rotation is applied to all simplex prototypes to preserve their relative separation. These prototypes are kept fixed during representation training.

Based on these fixed prototypes, the prototype-guided loss for sample $\mathbf{X}_i$ is defined as follows:

\begin{equation}
    \mathcal{L}_{p}^{(i)}
    =
    -\log
    \frac{
    \exp(\mathbf{z}_i^{T}\mathbf{p}_{y_i}/\tau)
    }{
    \exp(\mathbf{z}_i^{T}\mathbf{p}_{y_i}/\tau)
    +
    \sum_{j \ne i}
    \exp(\mathbf{z}_i^{T}\mathbf{z}_{j}/\tau)
    },
\end{equation}
where $\mathbf{p}_{y_i}$ denotes the prototype corresponding to the label $y_i$.

\subsection*{3) The Overall Loss Function}

Inspired by the prototype gating strategy for imbalanced supervised contrastive learning~\cite{mildenberger2025tale}, FreMGP uses a prototype gate to decide whether the prototype-guided loss should be applied to each sample. This avoids imposing overly strict constraints on samples that are already close to their class prototypes. Specifically, the gate is defined as follows:
\begin{equation}
    g_i =
    \begin{cases}
    1, & \mathbf{z}_i^{T}\mathbf{p}_{y_i} < \gamma, \\
    0, & \mathbf{z}_i^{T}\mathbf{p}_{y_i} \ge \gamma,
    \end{cases}
\end{equation}
where $\gamma$ is a similarity threshold. When $g_i=1$, the embedding $\mathbf{z}_i$ is not close enough to its class prototype, and the prototype-guided loss is activated. When $g_i=0$, the embedding $\mathbf{z}_i$ is already close to its class prototype, so the additional prototype constraint is not applied. 

Following prior work that selectively applies prototype constraints in supervised contrastive learning for imbalanced data~\cite{mildenberger2025tale}, FreMGP uses the supervised contrastive loss as the basic representation learning objective and selectively adds the prototype-guided loss according to the prototype gate. The final representation learning objective is defined as follows:
\begin{equation}
    \mathcal{L}_{rep}
    =
    \frac{1}{B}
    \sum_{i=1}^{B}
    \left(
    \mathcal{L}_{sup}^{(i)}
    +
     g_i \mathcal{L}_{p}^{(i)}
    \right),
\end{equation}
where $B$ is the number of embeddings in a batch.

Due to the page limit, other implementation details, including the frequency-domain MLP encoder inspired by frequency-domain MLPs for time-series forecasting (FreTS)~\cite{yi2023frequencydomain} and the projector architecture, are provided in Appendix B.1.

\subsection{Frequency-domain Contrastive Representation-Guided MTGP Oversampling}
\subsection*{1) Individual Representation}
In FreMGP, each individual contains multiple trees, each of which represents one synthetic frequency-domain time-series sample.
Let $N_c$ denote the number of training samples in the minority class $c$, and let $N_{\max}$ denote the number of training samples in the majority class. The number of synthetic samples required for class $c$ is then given by $M_c=N_{\max}-N_c$. Therefore, in FreMGP, each individual consists of $M_c$ strongly-typed trees, denoted as $\mathcal{T}_c=\{T_1,T_2,\ldots,T_{M_c}\}$, where each tree represents one synthetic frequency-domain sample. As illustrated in Fig.~\ref{fig:stgp}, a tree consists of four layers: \textit{Input}, \textit{Spectral Transform}, \textit{Spectral Fusion}, and \textit{Output}. The nodes in the \textit{Input} layer are selected from the terminal set, while the nodes in the \textit{Spectral Transform} and \textit{Spectral Fusion} layers are selected from the function set. The \textit{Output} layer returns the final synthetic frequency-domain sample. The terminal set and function set are introduced as follows:

\begin{figure}
    \centering
    \includegraphics[width=1\linewidth]{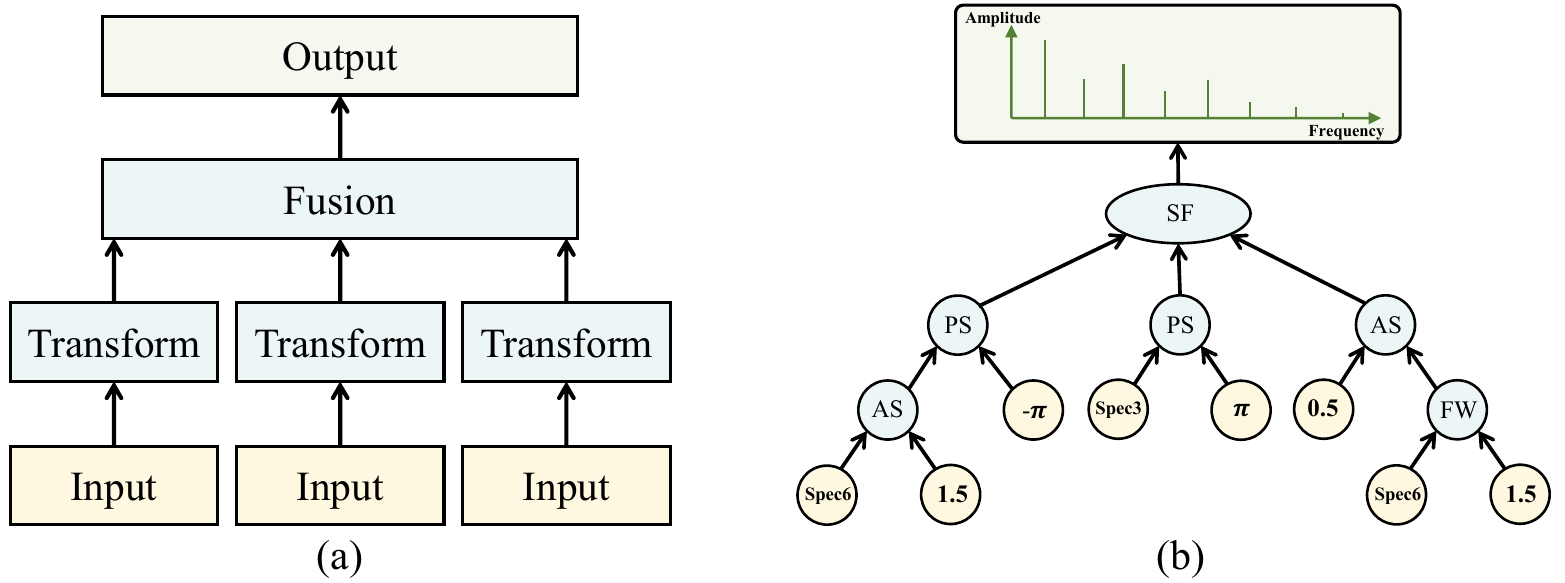}
    \caption{Illustration of the STGP tree representation in FreMGP. 
    (a) Layered structure of an STGP tree. 
    (b) Example of an STGP tree.}
    \label{fig:stgp}
\end{figure}

\paragraph{Terminal Set}


\begin{table*}[t]
\centering
\caption{The terminal set and function set used in FreMGP}
\label{tab:primitive_set}
\setlength{\tabcolsep}{8pt}
\renewcommand{\arraystretch}{1.08}
\resizebox{0.95\textwidth}{!}{
\begin{tabular}{lcc@{\hspace{1.2cm}}|lcc}
\hline
\multicolumn{3}{c|}{Terminal Set} 
& \multicolumn{3}{c}{Function Set} \\
\hline
Name & Type & Description 
& Name & Input Type & Output Type \\
\hline
Spectrum terminal & $\mathsf{S}$ & Segmented training spectrum 
&AS & $[\mathsf{S},\mathsf{A}]$ & $\mathsf{S}$ \\
ERC-AS & $\mathsf{A}$ & Amplitude scaling coefficient 
&PS & $[\mathsf{S},\mathsf{B}]$ & $\mathsf{S}$ \\
ERC-PS & $\mathsf{B}$ & Phase shifting coefficient 
& FW & $[\mathsf{S},\mathsf{C}]$ & $\mathsf{S}$ \\
ERC-FW & $\mathsf{C}$ & Frequency warping coefficient 
& SF & $[\mathsf{S},\mathsf{S},\mathsf{S}]$ & $\mathsf{O}$ \\
\hline
\end{tabular}
}
\begin{flushleft}
\footnotesize
\textbf{Note}: $\mathsf{S}$, $\mathsf{A}$, $\mathsf{B}$, $\mathsf{C}$, and $\mathsf{O}$ denote the spectrum type, amplitude scaling coefficient type, phase shifting coefficient type, frequency warping coefficient type, and output type, respectively. ERC-AS, ERC-PS, and ERC-FW are sampled from $(0,2]$, $[-\pi,\pi]$, and $[0.5,2.0]$, respectively.
\end{flushleft}
\end{table*}

The terminal set contains spectrum terminals and ephemeral random constants (ERCs). The terminal set is summarized in Table~\ref{tab:primitive_set}. The spectrum terminals are constructed from the frequency-domain samples of the training set.


To construct spectrum terminals, FreMGP decomposes each training spectrum into several frequency-band components. Specifically, each spectrum sample $\mathbf{X}_i$ is divided into three consecutive frequency segments, corresponding to the low-frequency, mid-frequency, and high-frequency segments. The first two segments have length $\lfloor F/3 \rfloor$, where $F$ denotes the dimension of the frequency-domain data obtained by the DFT in equation~\eqref{eq:dft}. The last segment contains the remaining frequency coefficients.
For each spectrum sample $\mathbf{X}_i$, we construct its $v$-th spectrum terminal by retaining only the coefficients in the $v$-th frequency segment and setting all coefficients in the other segments to zero. In this way, the original training spectra are decomposed into local spectral components. These components serve as spectrum terminals and can be recombined by GP functions to generate new spectra. Each training spectrum produces three spectrum terminals, and the total number of spectrum terminals is $N \times 3$, where $N$ denotes the total number of training samples across all classes. Each spectrum terminal is a complex-valued array of type $\mathsf{S}$.

Besides spectrum terminals, FreMGP uses three types of ERCs as scalar parameters for spectral transformations. Specifically, ERC-AS, ERC-PS, and ERC-FW correspond to amplitude scaling, phase shifting, and frequency warping, respectively.

\paragraph{Function Set}

FreMGP defines four spectral functions: amplitude scaling (AS), phase shifting (PS), frequency warping (FW), and spectral fusion (SF), summarized in Table~\ref{tab:primitive_set}. The AS function adjusts the magnitude of a spectrum, the PS function modifies its phase, the FW function performs a non-linear transformation along the frequency axis, and the SF function fuses multiple transformed spectral branches into the final output spectrum. These functions are defined as follows:
\begin{equation}
\begin{aligned}
    \mathrm{AS}(\mathbf{S},a) &= a\mathbf{S},\\
    \mathrm{PS}(\mathbf{S},b) &= \mathbf{S}\odot \exp(\varsigma b),\\
    \mathrm{FW}(\mathbf{S},c_w)_k &= \mathcal{I}_{\mathbf{S}}(\omega_k^{c_w}),\\
    \mathrm{SF}(\mathbf{S}_1,\mathbf{S}_2,\mathbf{S}_3)
    &= \sum_{i=1}^{3}\mathbf{S}_i,
\end{aligned}
\end{equation}
where $\mathbf{S}\in\mathbb{C}^{F}$ denotes a spectrum, $a\in\mathsf{A}$ is an amplitude scaling coefficient, $b\in\mathsf{B}$ is a phase shifting coefficient, $c_w\in\mathsf{C}$ is a frequency warping coefficient, $\varsigma$ is the imaginary unit satisfying $\varsigma^2=-1$, $\odot$ denotes element-wise multiplication, $\omega_k$ denotes the normalized frequency location at frequency index $k$, and $\mathcal{I}_{\mathbf{S}}(\cdot)$ denotes a linear interpolation on $\mathbf{S}$ for obtaining the spectral value at the warped frequency location.
In each tree, SF serves as the root node and corresponds to the fusion layer in Fig.~\ref{fig:stgp}(a), where different spectral branches are fused. The other functions, including AS, PS, and FW, are used in the transform layer for constructing spectral branches.

\subsection*{2) Target Sample Assignment and Two-Stage Evolutionary Search}

Before the evolutionary oversampling process, FreMGP assigns one target sample to each of the $M_c$ trees in an individual. These target samples are selected from the original samples of the minority class $c$. Specifically, the original samples of minority class $c$ are first ranked in ascending order of their distances to the class center. These ranked samples are then assigned to the trees until all the $M_c$ trees in an individual have been assigned their corresponding target samples. Note that this work focuses only on the imbalance case where the imbalance ratio (IR) is greater than 2. In that case, $M_c$ is greater than $N_c$ ($N_c$ is the number of training samples in the minority class $c$), and all the $N_c$ original minority-class samples must be used at least once. Therefore, the assignment process starts from the beginning of the ranked list and repeats cyclically until all $M_c$ trees have their own target samples. Each tree then uses its assigned target sample as a reference to generate one synthetic sample. Note that multiple trees in the same individual may share the same target minority-class sample for generation.

In FreMGP, the evolutionary oversampling process consists of two stages. Stage I is called \textit{Representation Proximity Search}, which aims to drive generated samples toward the target minority-class region in the learned representation space. In this stage, the evolutionary search focuses mainly on improving the representation proximity between generated samples and their corresponding target minority-class samples. Stage II is called \textit{Local Distribution Expansion Search}. This stage further improves the generated sample set by considering both similarity and diversity. Specifically, generated samples are encouraged to remain close to the target minority-class region while expanding in different directions around the target samples. Therefore, Stage II promotes diversity while keeping generated samples close to the target minority-class region.

FreMGP employs an adaptive criterion to determine when to transition from Stage I to Stage II. Let  $\bar{\mathcal{F}}_{\mathrm{I}}^{(e)}$ denote the average fitness values of  the population at generation $e$ in Stage I. The proximity threshold $\delta$ is adaptively determined according to the average fitness of the initial population in Stage I, defined as:
\begin{equation}
\delta =
\bar{\mathcal{F}}_{\mathrm{I}}^{(0)}
+
\lambda
\left(
1-\bar{\mathcal{F}}_{\mathrm{I}}^{(0)}
\right),
\end{equation}
where $\bar{\mathcal{F}}_{\mathrm{I}}^{(0)}$ is the average fitness 
of the initial population in Stage I, and $\lambda$ controls how far the threshold moves from the initial fitness level toward the ideal fitness value of 1.

If the average fitness of Stage I remains above the threshold $\delta$ for five consecutive generations, the population is considered to have reached a reliable minority-class region in the learned representation space. The search then switches from Stage I to Stage II.

\subsection*{3) Fitness Functions in the Two-Stage Search}
After the representation space is learned, the encoder $E_{\theta}(\cdot)$ is fixed and used to evaluate generated samples during evolutionary search. For the $m$-th generated frequency-domain sample $\widehat{\mathbf{X}}_{c}^{(m)}$ of class $c$, its representation is computed as:
\begin{equation}
    \widehat{\mathbf{h}}_{c}^{(m)}
    =
    E_{\theta}
    \left(
    \widehat{\mathbf{X}}_{c}^{(m)}
    \right).
\end{equation}

\paragraph*{a) The fitness function in Stage I}
This fitness function is designed to drive the \textit{Representation Proximity Search}, guiding generated samples toward their assigned target samples in the learned representation space. For the $m$-th tree in the current individual, its representation proximity score is defined as:
\begin{equation}
\label{eq:tree_score}
    s_c^{(m)}
    =
    \Phi\left(
    \left\|
    \widehat{\mathbf{h}}_{c}^{(m)}
    -
    \mathbf{h}_{c}^{(m)}
    \right\|_{2};
    \rho
    \right),
\end{equation}
where $\mathbf{h}_{c}^{(m)}$ is the original representation of the assigned target sample, $\widehat{\mathbf{h}}_{c}^{(m)}$ is the representation of the generated sample produced by this tree, $\rho$ is an adaptive scale parameter computed as the median pairwise distance among target minority-class representations, and $\Phi(d;\rho)$ is the Gaussian mapping function defined as:

\begin{equation}
\Phi(d;\rho)
=
\exp
\left(
-
\frac{d^2}{2\rho^2}
\right)
\label{eq:gaussian_mapping}
\end{equation}

The major advantage of using Gaussian mapping is that its decay behavior is explicitly controlled by the scale parameter $\rho$, instead of using a fixed threshold to judge whether a generated sample is close enough. 

Based on the tree-level representation proximity scores, the fitness function for Stage I is defined as:
\begin{equation}
\label{eq:stage1_fitness}
    \mathcal{F}_{\mathrm{I}}
    =
    \frac{1}{M_c}
    \sum_{m=1}^{M_c} s_c^{(m)}.
\end{equation}

According to this fitness function, when the representation of a generated sample produced by a tree is close to the representation of its assigned target sample, the obtained score $s_c^{(m)}$ approaches 1; otherwise, the score decreases as their distance increases. Thus, by averaging the proximity scores over all the $M_c$ generated samples, a larger $\mathcal{F}_{\mathrm{I}}$ score indicates that the samples generated by the trees are closer to their corresponding target minority-class samples in the learned representation space. Therefore, $\mathcal{F}_{\mathrm{I}}$ encourages an individual to maintain representation proximity to the target minority-class region.

\paragraph*{b) The fitness function in Stage II}

Although the fitness function in Stage I encourages generated samples to approach their assigned target samples, considering only the averaged representation proximity fitness may lead to a sample set with low diversity. This is because, based on the target sample assignment method, multiple trees in an individual may share the same target minority-class sample. To address the above issue, we categorize trees (i.e., generated samples) sharing the same target sample into a group, and evaluate the fitness of an individual in a group-wise manner in Stage II. Note that, within the same group, some generated samples may be very close to their target samples in the learned representation space while others remain relatively far away. This can still result in a high average fitness. Therefore, Stage II evaluates groups of generated samples by considering both radial consistency and angular diversity.

The radial consistency score is designed to measure whether generated samples within the same group are close to their shared target sample in the learned representation space while maintaining a stable expansion radius. Specifically, for the $q$-th group, we first compute the Euclidean distances between the representations of all generated samples in this group and the representation of their shared target sample. The group-level radial consistency score is defined as:
\begin{equation}
    S_{\mathrm{rad}}^{(q)}
    =
    \Phi\left(\bar{\Delta}_q;\rho\right)
    \cdot
    \Phi\left(\sigma_{\Delta,q};\rho\right),
\end{equation}
where $\Phi(\cdot;\rho)$ is the Gaussian proximity mapping defined in
equation~\eqref{eq:gaussian_mapping}, and $\rho$ is the adaptive scale parameter defined in Stage I. 
$\bar{\Delta}_q$ and $\sigma_{\Delta,q}$ denote the mean and standard deviation of the radial distances between the representations of the generated samples in the $q$-th group and the representation of their shared target sample:
\begin{equation}
    \bar{\Delta}_q
    =
    \frac{1}{K_q}
    \sum_{k=1}^{K_q}
    \left\|
    \widehat{\mathbf{h}}_{q,k}
    -
    \mathbf{h}_q
    \right\|_2 ,
\end{equation}
\begin{equation}
    \sigma_{\Delta,q}
    =
    \sqrt{
    \frac{1}{K_q}
    \sum_{k=1}^{K_q}
    \left(
    \left\|
    \widehat{\mathbf{h}}_{q,k}
    -
    \mathbf{h}_q
    \right\|_2
    -
    \bar{\Delta}_q
    \right)^2
    },
\end{equation}
where $\widehat{\mathbf{h}}_{q,k}$ denotes the representation of the $k$-th generated sample in the $q$-th group, $\mathbf{h}_q$ denotes the shared target representation of this group, and $K_q$ is the number of generated samples in this group.

\begin{figure*}
\centering
  \includegraphics[width=1.0\textwidth]{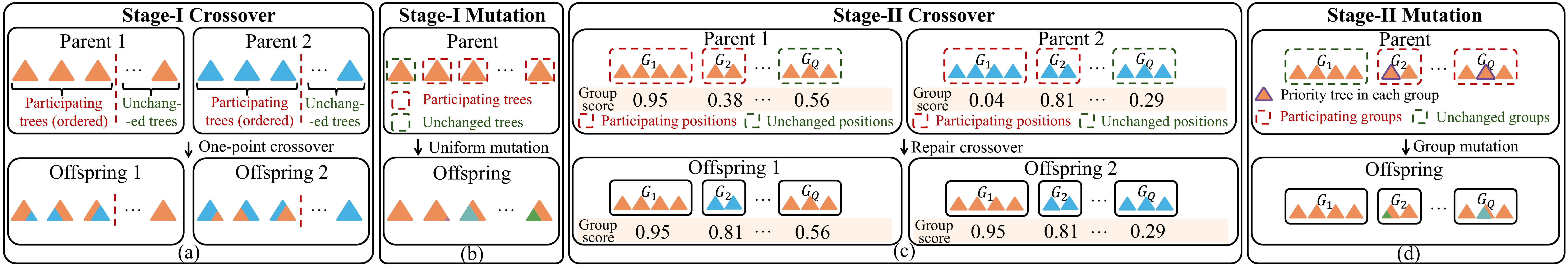}
  \caption{Illustration of the stage-specific genetic operators in FreMGP. 
(a) Stage-I one-point subtree crossover with trees as the basic operation units. 
(b) Stage-I uniform subtree mutation with trees as the basic operation units. 
(c) Stage-II group-level repair-style crossover, where higher-scoring groups replace lower-scoring groups at the same group positions. 
(d) Stage-II group-level mutation, where participating groups are refined by mutating their priority trees.}
\label{fig:stage_specific_operators}
  \label{fig:crossover_mutation}
\end{figure*}

This score $S_{\mathrm{rad}}^{(q)}$ measures radial consistency directly. 
The first factor evaluates the average proximity of generated samples to the shared target sample in the learned representation space, while the second evaluates the consistency of their radial distances. The product in $S_{\mathrm{rad}}^{(q)}$ assigns a high score only when both requirements are satisfied.

Although radial consistency encourages generated samples close to the shared target sample in the learned representation space with stable radial distances, it does not explicitly encourage them to move toward different directions. Therefore, FreMGP further introduces an angular diversity score. For each generated sample, we compute its direction of movement by normalizing the vector from the corresponding target representation to the representation of the generated sample:
\begin{equation}
    \boldsymbol{\xi}_{q,k}
    =
    \frac{
    \widehat{\mathbf{h}}_{q,k}-\mathbf{h}_q
    }{
    \left\|
    \widehat{\mathbf{h}}_{q,k}-\mathbf{h}_q
    \right\|_2+\epsilon
    },
\end{equation}
where $\widehat{\mathbf{h}}_{q,k}$ denotes the representation of the $k$-th generated sample in the $q$-th group, $\mathbf{h}_q$ denotes the shared target representation of this group, and $\epsilon$ is a small constant for numerical stability.

Therefore, the group-level angular diversity score is defined as:
\begin{equation}
    S_{\mathrm{ang}}^{(q)}
    =
    \frac{2}{K_q(K_q-1)}
    \sum_{1\leq k<s\leq K_q}
    \frac{
    1-
    \boldsymbol{\xi}_{q,k}^{\top}
    \boldsymbol{\xi}_{q,s}
    }{2},
\end{equation}
where $K_q$ is the number of generated samples in the $q$-th group. This score averages the pairwise directional differences among generated samples in the same group. A higher $S_{\mathrm{ang}}^{(q)}$ indicates that the generated
samples move along more diverse directions around the shared target representation. 
Note that if a group contains only one generated sample, there is no pairwise directional difference to compute, and its angular diversity score is set to the radial consistency score.

The total score of group $q$ is defined as:
\begin{equation}
    G_q
    =
    \alpha S_{\mathrm{rad}}^{(q)}
    +
    (1-\alpha)S_{\mathrm{ang}}^{(q)},
\label{eq:stage2_group_score}
\end{equation}
where $\alpha$ is a weight to balance between radial consistency and angular diversity.

Therefore, the fitness function in Stage II is defined as: 
\begin{equation}
    \mathcal{F}_{\mathrm{II}}
    =
    \frac{1}{Q}
    \sum_{q=1}^{Q} G_q,
\label{eq:stage2_fitness}
\end{equation}
where $Q$ is the number of groups. 

\subsection*{4) Genetic Operators}

We design different genetic operators for evolution in different search stages. Specifically, genetic operations are conducted on individual trees in Stage I, but at the group level in Stage II.
\paragraph*{a) The genetic operators in Stage I}

To maintain consistency with the per-tree evaluation in Stage I, FreMGP conducts genetic operations at the tree level. Each tree in an individual has its own representation proximity score, calculated by equation \eqref{eq:tree_score}, which can be used to calculate its participation probability for crossover or mutation. The participation probability of tree $T_m$ is defined as:

\begin{equation}    
p_m    
=    
\frac{    
s_{\max}-s_c^{(m)}    
}{    
\sum_{j=1}^{M_c}    
\left(    
s_{\max}-s_c^{(j)}    
\right)    
},
\end{equation}
where $s_{\max}=\max_{1\leq j\leq M_c}s_c^{(j)}$ is the maximum tree-level representation proximity score within the current individual. According to this equation, lower-scoring trees receive higher probabilities.

Within each parent individual, half of the trees are selected probabilistically according to their participation probabilities to undergo crossover or mutation, while the other half remain unchanged. As shown in Fig.~\ref{fig:stage_specific_operators}(a), the selected trees are then paired according to their participation order, and one-point sub-tree crossover is performed on each pair. As shown in Fig.~\ref{fig:stage_specific_operators}(b), the mutation operator uses the same per-tree participation mechanism and applies uniform sub-tree mutation to breed.

\paragraph*{b) The genetic operators in Stage II}
In Stage II, each individual is evaluated based on a group of generated samples associated with the same target sample. Therefore, genetic operations at this stage are also performed at the group level. 

As shown in Fig.~\ref{fig:stage_specific_operators}(c), the crossover operator adopts a group-level strategy. For the same group position in two parent individuals, the group with the higher score (calculated by equation (\ref{eq:stage2_group_score})) in one parent is copied to replace the lower-scoring group at the same position in the other parent. In Stage II, only $25\%$ of the group positions are allowed to perform crossover. For two parent individuals, each group position is assigned a crossover participation probability according to the score difference between the two groups at the same position. The participation probability of the $q$-th group position is defined as:
\begin{equation}
    p_q^{\mathrm{cx}}
    =
    \frac{
    \left|G_q^{(1)}-G_q^{(2)}\right|
    }{
    \sum_{r=1}^{Q}
    \left|G_r^{(1)}-G_r^{(2)}\right|
    },
\end{equation}
where $G_q^{(1)}$ and $G_q^{(2)}$ denote the scores of the $q$-th group in the two parent individuals, respectively, and $Q$ is the total number of groups. This probability is proportional to the score difference between the two parent
groups at the same position. Therefore, group positions with larger quality gaps are more likely to participate in crossover.

As shown in Fig.~\ref{fig:stage_specific_operators}(d), mutation in Stage II is also performed with groups. To give low-quality groups more opportunities for refinement, groups with lower scores are assigned higher mutation participation probabilities. The mutation participation probability of group $q$ is defined as:
\begin{equation}    
p_q^{\mathrm{mut}}    
=    
\frac{    
G_{\max}-G_q    
}{    
\sum_{r=1}^{Q}    
\left(    
G_{\max}-G_r    
\right)    
},
\end{equation}
where $G_q$ denotes the score of the $q$-th group, $Q$ is the total number of groups, and $G_{\max}=\max_{1\le r\le Q}G_r$ is the maximum group score in the current individual. This probability is inversely related to the group score, so groups with lower $G_q$ values are more likely to participate in mutation.

The same as the crossover operator in Stage II, only $25\%$ of the groups in an individual are allowed to mutate according to their mutation participation probability. Within each mutation group, FreMGP further prioritizes a single tree to be mutated. The priority of the $k$-th tree in the $q$-th group is computed using a leave-one-out strategy, i.e., a tree is assigned a higher priority if its removal leads to a larger increase in the group score. This is defined as:
\begin{equation}    
\eta_{q,k}    
=    
\alpha    
\left[    
S_{\mathrm{rad}}^{(q,-k)}    
-    
S_{\mathrm{rad}}^{(q)}    
\right]_{+}    
+    
(1-\alpha)    
\left[    
S_{\mathrm{ang}}^{(q,-k)}    
-    
S_{\mathrm{ang}}^{(q)}    
\right]_{+},
\end{equation}
where $S_{\mathrm{rad}}^{(q,-k)}$ and $S_{\mathrm{ang}}^{(q,-k)}$ denote the radial consistency and angular diversity scores after removing the $k$-th tree from the $q$-th group, and $[\cdot]_{+}$ denotes the positive part. Here, $\alpha$ is the same weighting parameter as in equation~(\ref{eq:stage2_group_score}). A larger $\eta_{q,k}$ indicates that removing this tree improves the group quality more, so this tree is assigned a higher priority.

\paragraph*{c) The elitism strategies}

To retain high-quality trees, groups, and individuals during evolution, FreMGP uses elitism in both stages. In Stage I, the tree-level elitism is performed. At each tree position index, FreMGP compares the corresponding trees from different individuals and selects the one with the highest representation proximity score (equation (\ref{eq:tree_score})). This process continues until every tree position index has identified its best-performing tree. Afterwards, the best-performing trees identified at each tree position index are recombined to form new elite individuals. This tree-level elitism preserves the trees that have already been able to generate high-quality samples in Stage I.

In Stage II, the group-level elitism is performed. At each group index, FreMGP compares scores of groups (calculated by the equation (\ref{eq:stage2_group_score})) from different individuals and selects the better-performing group. This process continues until every group position index has identified its best-performing group. Afterwards, the best-performing groups identified at each group index are recombined to form new elite individuals.
In addition, in Stage II, FreMGP also directly retains the individual with the highest fitness value. This allows preserving both locally well-performing groups and globally well-performing individuals.

\section{Experimental Design}

\begin{table}
\caption{Datasets in the experiments}
\label{tab:datasets}
\centering
\begin{tabular}{ccccc}
\hline
Dataset & No.sample & Class & IR & Length  \\ \hline
MixedShapesSmallTrain & 60 & 5 & 2.22 & 1024 \\
SonyAIBORobotSurface1 & 20 & 2 & 2.33 & 70 \\
TwoLeadECG & 17 & 2 & 2.4 & 82 \\
Ham & 71 & 2 & 4.07 & 431  \\
ECG200 & 80 & 2 & 6.27 & 96  \\
MiddlePhalanxOutlineCorrect & 438 & 2 & 7.76 & 80  \\
SyntheticControl & 90 & 6 & 10 & 60  \\
DistalPhalanxOutlineAgeGroup & 377 & 3 & 10.71 & 80  \\
SmoothSubspace & 68 & 3 & 12.5 & 15 \\
PowerCons & 97 & 2 & 12.86 & 144  \\
SwedishLeaf & 140 & 15 & 14 & 128  \\
Strawberry & 414 & 2 & 19.7 & 235  \\
StarLightCurves & 612 & 3 & 40.93 & 1024 \\
Wafer & 917 & 2 & 64.42 & 152 \\
\hline
\end{tabular}

\vspace{0.5ex}
\begin{minipage}{0.95\linewidth}
\footnotesize
\textbf{Note}: For multi-class datasets, IR is the ratio of the number of samples in the largest class to the number in the smallest class.
\end{minipage}
\end{table}

\subsection{Datasets}

In our experiments, we use 14 time-series datasets from the UCR Archive\footnote{These datasets are available at:\href{https://www.cs.ucr.edu/~eamonn/time_series_data/}{\url{https://www.cs.ucr.edu/~eamonn/time_series_data/}}}. These datasets cover diverse domains, including healthcare, food analysis, industrial monitoring, robotics, and other real-world scenarios. Note that the UCR datasets provide predefined training/test splits.  Following~\cite{zhao2022tsmote}, we retain the original test sets and perform stratified sampling on the training sets to produce new imbalanced datasets that cover a wider range of IR values. 
Table~\ref{tab:datasets} reports the number of training samples after stratified sampling, the number of classes, the IR, and the time-series length. The IR values range from 2.22 to 64.42. 

\subsection{Baseline Methods}

The baseline methods cover representative oversampling methods, including SMOTE~\cite{chawla2002smote}, Borderline-SMOTE1 (abbreviated as BL-SMOTE)~\cite{han2005borderline}, ADASYN~\cite{he2008adasyn}, SVM-SMOTE~\cite{nguyen2011borderline}, and KMeans-SMOTE (abbreviated as KM-SMOTE)~\cite{douzas2018improving}. We also include time-series-specific oversampling methods, including T-SMOTE~\cite{zhao2022tsmote}, INOS~\cite{cao2013integrated}, and OHIT~\cite{zhu2022minority}. Furthermore, deep-learning-based generative methods, including CSMOTE~\cite{liu2021csmote}, BFGAN~\cite{lee2022boundary}, CFAMG~\cite{wang2025mitigating}, and ImagenFew~\cite{gonen2025datascarcity}, are included as baseline method. Finally, Evo-TFS~\cite{pei2026evotfs} is also compared since it is the recent GP-based evolutionary oversampling method. Overall, these baseline methods span a wide range of oversampling strategies, allowing a comprehensive comparison with FreMGP. Detailed descriptions on these baseline methods are provided in Appendix C.

We employ three types of classifiers, namely LSTM networks \cite{hochreiter1997long}, Transformer \cite{vaswani2017attention}, and Mamba \cite{gu2023mamba}, to evaluate the quality of the generated time-series data. Each classifier is trained on the rebalanced training set generated by FreMGP or a baseline method, and then evaluated on the same original test set.


\subsection{Parameter Settings}

\begin{table}
\centering
\caption{Parameter settings}
\label{tab:parameter_settings}
\begin{tabular}{>{\centering\arraybackslash}m{1.4cm}cc}
\hline
Module & Parameter & Value \\
\hline
\multirow{4}{*}{\makecell[c]{Representation\\learning}}
& Encoder output dimension $d_h$ & 256 \\
& Projection output dimension $d_z$ & 128 \\
& Temperature $\tau$ & 0.1 \\
& Prototype gate threshold $\gamma$ & 0.5 \\
\hline
\multirow{9}{*}{\makecell[c]{Evolutionary\\oversampling}}
& Initialization method & Ramped half-and-half \\
& Number of generations & 100 \\
& Population size & 128 \\
& Tournament size & 3 \\
& Crossover rate & 0.8 \\
& Mutation rate & 0.2 \\
& Maximum tree depth & 10 \\
& Stage transition coefficient $\lambda$ & 0.7 \\
& $\alpha$ & 0.5 \\
\hline
\end{tabular}
\end{table}

Table~\ref{tab:parameter_settings} presents the parameter configurations of the proposed FreMGP method. 
Following common practices in contrastive representation learning \cite{chen2020simple,khosla2020supervised}, the encoder output dimension $d_h$ and the projection dimension $d_z$ are set to 256 and 128, respectively, to provide a practical balance between representation capacity and computational cost. 
The temperature parameter $\tau$ is set to 0.1, which is a commonly used setting in supervised contrastive learning \cite{khosla2020supervised}. In addition, inspired by the prototype-based gating strategy for imbalanced supervised contrastive learning \cite{mildenberger2025tale}, the prototype gate threshold $\gamma$ is set to 0.5 to selectively activate the prototype-guided loss. Other implementation details of the representation learning module, such as the optimizer and training epochs, are provided in the Appendix B.2.

\begin{table*}[t]
\centering
\caption{Overall performance comparison of FreMGP and baseline methods under different classifiers and evaluation metrics}
\label{tab:classifier_comparison}
\resizebox{\textwidth}{!}{
\begin{tabular}{lccccccccc}
\toprule
\multirow{2}{*}{Methods} 
& \multicolumn{3}{c}{LSTM} 
& \multicolumn{3}{c}{Transformer} 
& \multicolumn{3}{c}{Mamba} \\
\cmidrule(lr){2-4} \cmidrule(lr){5-7} \cmidrule(lr){8-10}
& F1-Score & G-Mean & AUC 
& F1-Score & G-Mean & AUC
& F1-Score & G-Mean & AUC \\
\midrule
None          & 0.4685 & 0.2254 & 0.7358
              & 0.6042 & 0.4641 & 0.8505
              & 0.6721 & 0.5614 & 0.8930 \\
\midrule
SMOTE         & 0.6899 & 0.6739 & 0.8231
              & 0.7163 & 0.6882 & 0.8666
              & 0.7540 & 0.7243 & 0.8952 \\

BL-SMOTE      & 0.6691 & 0.6261 & 0.8218
              & 0.7109 & 0.6767 & 0.8626
              & 0.7404 & 0.7023 & 0.8787 \\

ADASYN        & 0.6849 & 0.6671 & 0.8232
              & 0.7292 & 0.7041 & 0.8769
              & 0.7517 & 0.7206 & 0.8975 \\

SVM-SMOTE     & 0.6668 & 0.6346 & 0.8133
              & 0.7106 & 0.6731 & 0.8729
              & 0.7496 & 0.7132 & 0.8889 \\

KM-SMOTE      & 0.6830 & 0.6524 & 0.8226
              & 0.7049 & 0.6616 & 0.8589
              & 0.7434 & 0.7025 & 0.8844 \\

\midrule
T-SMOTE       & 0.5549 & 0.3984 & 0.7818
              & 0.6663 & 0.5792 & 0.8436
              & 0.7561 & 0.7196 & 0.8876 \\

INOS          & 0.6791 & 0.6396 & 0.8284
              & \underline{0.7340} & \underline{0.7090} & \underline{0.8859}
              & \underline{0.7734} & \underline{0.7493} & \underline{0.9094} \\

OHIT          & \underline{0.7043} & \underline{0.6871} & \underline{0.8309}
              & 0.7266 & 0.6973 & 0.8794
              & 0.7512 & 0.7040 & 0.9011 \\

CSMOTE       & 0.6696 & 0.6506 & 0.8278
              & 0.7101 & 0.6802 & 0.8811
              & 0.7389 & 0.6938 & 0.8998 \\

BFGAN        & 0.5222 & 0.3544 & 0.7944
              & 0.6306 & 0.4790 & 0.8443
              & 0.6903 & 0.5996 & 0.8829 \\

CFAMG         & 0.4184 & 0.2561 & 0.7038
              & 0.4530 & 0.2726 & 0.7199
              & 0.4036 & 0.1915 & 0.7341 \\

ImagenFew     & 0.5274 & 0.3131 & 0.7590
              & 0.6013 & 0.4648 & 0.8267
              & 0.6916 & 0.5878 & 0.8439 \\

Evo-TFS       & 0.6289 & 0.5515 & 0.7995
              & 0.6774 & 0.5975 & 0.8647
              & 0.7299 & 0.6729 & 0.8898 \\
\midrule
FreMGP        & \textbf{0.7176} & \textbf{0.7049} & \textbf{0.8461}
              & \textbf{0.7621} & \textbf{0.7454} & \textbf{0.8971}
              & \textbf{0.8042} & \textbf{0.7878} & \textbf{0.9169} \\
\bottomrule
\end{tabular}
}
\end{table*}

For the evolutionary oversampling module, the population size is set to 128, and the maximum number of generations is set to 100 to ensure a sufficient number of evaluations. 
To prevent tree bloating during evolution, the maximum tree depth is set to 10. For the adaptive two-stage search, the stage transition coefficient $\lambda$ is set to 0.7. 
These settings ensure that the search enters Stage II only after the population has reached a reliable representation proximity level in Stage I. The weight $\alpha$ is set to 0.5 to equally balance radial consistency and angular diversity in the Stage-II fitness evaluation. FreMGP employs stage-specific elitism strategies. At Stage I, three new elite individuals are reconstructed based on the tree-level elitism; at Stage II, one new elite individuals are reconstructed based on the group-level elitism and two top individuals are directly retained from the last generation.


The representation learning module was implemented using PyTorch, while the multi-tree genetic programming was implemented based on the DEAP package \cite{fortin2012deap}. The conventional sampling baselines were implemented using the imbalanced-learn package \cite{lemaitre2017imbalanced}, and the other baseline methods were implemented based on their publicly available source code.

\section{Results and Analysis}

On each dataset, FreMGP is independently executed 30 times using 30 different random seeds. Three classifiers, LSTM, Transformer and Mamba, are trained on the training sets rebalanced by different sampling methods, and then their classification performances are evaluated on the same original test set, using F1-Score~\cite{sujon2025accuracy}, G-Mean~\cite{chen2024survey}, and AUC~\cite{richardson2024roc}. The Wilcoxon signed-rank test~\cite{demsar2006statistical} with a significance
level of 0.05 and the Holm--Bonferroni correction~\cite{holm1979simple} for multiple comparisons have been conducted to assess whether the performance difference between FreMGP and a baseline method is statistically significant.


\subsection{Overall Performance and Analysis}

\begin{table*}[htbp]
\centering
\caption{Statistical win/tie/loss comparison of FreMGP against baseline methods under different classifiers and evaluation metrics}
\label{tab:win_tie_loss}
\resizebox{\textwidth}{!}{
\begin{tabular}{lccccccccc}
\toprule
\multirow{2}{*}{Methods}
& \multicolumn{3}{c}{F1-Score}
& \multicolumn{3}{c}{G-Mean}
& \multicolumn{3}{c}{AUC} \\
\cmidrule(lr){2-4} \cmidrule(lr){5-7} \cmidrule(lr){8-10}
& LSTM (+/=/-) & Transformer (+/=/-) & Mamba (+/=/-)
& LSTM (+/=/-) & Transformer (+/=/-) & Mamba (+/=/-)
& LSTM (+/=/-) & Transformer (+/=/-) & Mamba (+/=/-) \\
\midrule
SMOTE         & 6/7/1  & 10/4/0 & 9/5/0
              & 6/8/0  & 11/3/0 & 11/3/0
              & 6/8/0  & 9/5/0  & 8/5/1 \\

BL-SMOTE      & 8/6/0  & 12/2/0 & 11/2/1
              & 8/6/0  & 12/2/0 & 12/1/1
              & 6/8/0  & 11/2/1 & 9/4/1 \\
              
ADASYN        & 6/7/1  & 10/2/2 & 9/5/0
              & 8/6/0  & 10/2/2 & 10/4/0
              & 6/8/0  & 7/6/1  & 6/5/3 \\
              
SVM-SMOTE     & 8/6/0  & 11/3/0 & 10/4/0
              & 9/5/0  & 13/1/0 & 10/4/0
              & 7/7/0  & 9/3/2  & 7/6/1 \\
              
KM-SMOTE      & 5/9/0  & 11/3/0 & 10/4/0
              & 7/7/0  & 12/2/0 & 10/4/0
              & 7/6/1  & 9/5/0  & 7/6/1 \\

\midrule
T-SMOTE       & 12/2/0 & 9/5/0  & 8/4/2
              & 12/2/0 & 9/5/0  & 9/3/2
              & 12/1/1 & 9/3/2  & 7/4/3 \\

INOS          & 7/7/0  & 7/6/1  & 9/4/1
              & 5/9/0  & 6/7/1  & 9/4/1
              & 6/7/1  & 7/5/2  & 5/5/4 \\

OHIT          & 4/7/3  & 7/7/0  & 7/6/1
              & 3/8/3  & 6/8/0  & 9/4/1
              & 3/8/3  & 7/5/2  & 7/3/4 \\

CSMOTE       & 9/3/2  & 11/3/0 & 11/3/0
              & 10/2/2 & 11/3/0 & 11/3/0
              & 9/4/1  & 8/3/3  & 8/1/5 \\

BFGAN        & 13/0/1 & 13/1/0 & 13/0/1
              & 12/1/1 & 13/1/0 & 13/1/0
              & 10/2/2 & 10/2/2 & 7/6/1 \\

CFAMG         & 14/0/0 & 13/0/1 & 14/0/0
              & 13/1/0 & 13/0/1 & 14/0/0
              & 14/0/0 & 13/0/1 & 13/0/1 \\

ImagenFew     & 12/0/2 & 11/1/2 & 11/2/1
              & 12/1/1 & 11/1/2 & 11/2/1
              & 12/2/0 & 11/1/2 & 9/4/1 \\

Evo-TFS       & 12/2/0 & 10/4/0 & 10/4/0
              & 11/3/0 & 12/2/0 & 11/3/0
              & 9/5/0  & 10/3/1 & 8/5/1 \\
\midrule
Total         & 116/56/10 & 135/41/6 & 132/43/7
              & 116/59/7  & 139/37/6 & 140/36/6
              & 107/66/9  & 120/43/19 & 101/54/27 \\
\bottomrule
\end{tabular}
}
\end{table*}

Table~\ref{tab:classifier_comparison} reports the average classification
performance of LSTM, Transformer, and Mamba classifiers using different sampling methods. The row ``None'' denotes the results obtained without using any oversampling method. The detailed results on each dataset are provided in the Appendix D.1. According to Table~\ref{tab:classifier_comparison}, compared with ``None'', most oversampling methods can assist the three classifiers to improve the classification performance. This suggests that rebalancing the training set is necessary for imbalanced time-series classification.
Among all the methods in the experiments, FreMGP achieves the best average results across all three classifiers and all three evaluation metrics. This demonstrates its effectiveness in rebalancing data to assist different classifiers to address the performance bias issue for ITSC tasks. 

Table~\ref{tab:win_tie_loss} reports the statistical comparison
between FreMGP and each baseline method under different classifiers and
evaluation metrics. Here, ``$+$'', ``$=$'', and ``$-$'' denote that FreMGP is significantly better than, not significantly different from, and significantly worse than the corresponding baseline method, respectively. 
For F1-Score, FreMGP achieves significantly better or statistically comparable results than the baseline methods in 172, 176, and 174 out of 182 total cases for the LSTM, Transformer, and Mamba classifiers, respectively. Similar trends can also be observed for G-Mean, where FreMGP 
achieves significantly better or statistically comparable results than the baseline methods in 175, 176, and 176 out of 182 total cases for the three classifiers, respectively.  For AUC, FreMGP also outperforms the baseline methods in most cases. 


\newcolumntype{C}[1]{>{\centering\arraybackslash}p{#1}}

\begin{table*}
\centering
\caption{Average ranking comparison of different oversampling methods under different classifiers}
\label{tab:average_rank}
\setlength{\tabcolsep}{3pt}
\renewcommand{\arraystretch}{0.9}
\resizebox{\textwidth}{!}{%
\begin{tabular}{l C{1.25cm} C{1.25cm} C{1.25cm} C{1.25cm} C{1.25cm} C{1.25cm} C{1.25cm} C{1.25cm} C{1.25cm}}
\toprule
\multirow{2}{*}{Methods}
& \multicolumn{3}{c}{LSTM}
& \multicolumn{3}{c}{Transformer}
& \multicolumn{3}{c}{Mamba} \\
\cmidrule(lr){2-4} \cmidrule(lr){5-7} \cmidrule(lr){8-10}
& F1-Score & G-Mean & AUC
& F1-Score & G-Mean & AUC
& F1-Score & G-Mean & AUC \\
\midrule
SMOTE         & 5.64 & 5.64 & 6.50
              & 8.18 & 7.68 & 9.07
              & 7.50 & 7.50 & 7.71 \\

BL-SMOTE      & 7.21 & 7.21 & 8.54
              & 7.89 & 7.39 & 8.46
              & 8.21 & 8.18 & 9.25 \\

ADASYN        & 6.36 & 6.07 & 7.54
              & 6.07 & 6.07 & 6.50
              & 6.79 & 7.00 & 7.07 \\

SVM-SMOTE     & 7.39 & 7.21 & 7.68
              & 8.04 & 8.07 & 6.36
              & 7.18 & 7.04 & 7.50 \\

KM-SMOTE      & 5.89 & 6.29 & 6.57
              & 8.39 & 8.75 & 8.29
              & 7.43 & 7.75 & 8.07 \\
\midrule
T-SMOTE       & 10.57 & 10.64 & 10.57
              & 8.00 & 8.21 & 8.50
              & 6.57 & 6.54 & 7.32 \\

INOS          & 6.39 & 6.00 & 6.43
              & \underline{4.96} & \underline{4.89} & \underline{5.39}
              & \underline{5.61} & \underline{5.43} & 6.57 \\

OHIT          & \underline{3.68} & \underline{3.75} & \underline{5.68}
              & 5.64 & 5.64 & 5.57
              & 6.39 & 6.68 & \underline{5.57} \\

CSMOTE        & 6.89 & 6.89 & 5.71
              & 7.82 & 7.36 & 7.46
              & 8.21 & 8.54 & 5.75 \\

BFGAN         & 11.64 & 11.36 & 9.25
              & 10.68 & 11.71 & 9.29
              & 10.82 & 10.89 & 9.32 \\

CFAMG         & 11.89 & 12.14 & 12.71
              & 12.86 & 12.36 & 13.54
              & 13.64 & 13.18 & 13.86 \\

ImagenFew     & 11.00 & 11.57 & 10.79
              & 10.64 & 9.96 & 10.57
              & 10.36 & 10.07 & 10.50 \\

Evo-TFS       & 8.96 & 8.96 & 9.14
              & 8.18 & 8.61 & 8.57
              & 8.46 & 8.32 & 9.21 \\
\midrule
FreMGP        & \textbf{3.61} & \textbf{3.00} & \textbf{4.00}
              & \textbf{2.43} & \textbf{2.29} & \textbf{3.82}
              & \textbf{2.79} & \textbf{2.57} & \textbf{4.96} \\
\bottomrule
\end{tabular}%
}
\end{table*}

\begin{figure*}
\centering
  \includegraphics[width=1.0\textwidth]{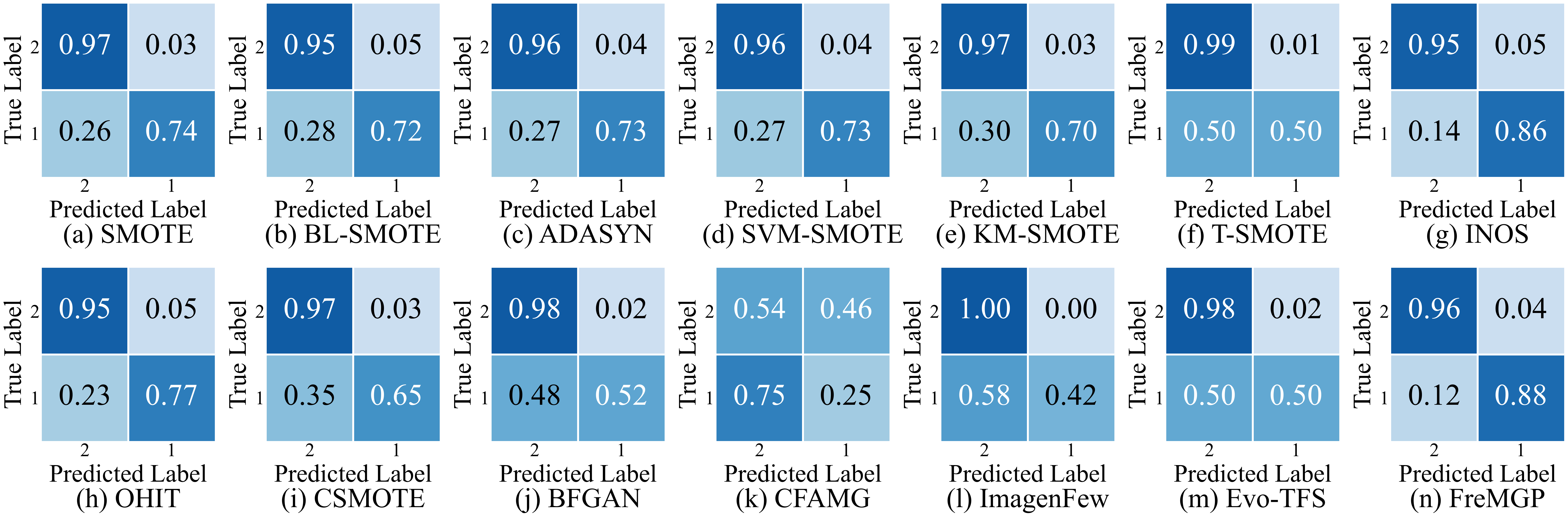}
  \caption{Comparison of confusion matrices for different oversampling methods on the Strawberry dataset with the Mamba classifier.
  }
  \label{fig:Matrix_Strawberry_mamba}
\end{figure*}

Table~\ref{tab:average_rank} reports the average rankings of different
oversampling methods over the 14 datasets under different classifiers and
evaluation metrics, where a smaller ranking value indicates better 
performance. Overall, FreMGP achieves the best average ranking across the three classifiers under all performance measures. This indicates that FreMGP is agnostic to specific classifiers and performs effectively across different time-series classifiers. Some baseline methods also perform effectively in certain cases. For example, OHIT ranks second under LSTM, and INOS performs competitively under Transformer and Mamba. 


\subsection{Minority-Class Recognition Analysis Based on Confusion Matrices}


To determine whether an oversampling method improves classification performance at the class level, we take the Strawberry dataset as an example, and confusion matrices are presented for qualitative analysis. Due to the page limit, only the confusion matrices obtained with the Mamba classifier are shown in the main paper, while the other results under the LSTM and Transformer classifiers are provided in  Appendix D.3.
As shown in Fig.~\ref{fig:Matrix_Strawberry_mamba}, the confusion matrices on the Strawberry dataset show that most methods obtain high accuracy for Class 2, while their performance on Class 1 varies more clearly. Therefore, the main difference among the compared methods lies in whether they can improve the accuracy on Class 1 without largely sacrificing the accuracy on Class 2.

After using FreMGP to rebalance data, Mamba achieves an accuracy of 0.88 on Class 1 while maintaining an accuracy of 0.96 on Class 2. Using INOS, Mamba also performs well on Class 1, with an accuracy of 0.86. However, compared with INOS, FreMGP assists Mamba in obtaining slightly higher accuracies for both classes. Several sampling methods, such as T-SMOTE, BFGAN, ImagenFew, and Evo-TFS, are less effective in assisting Mamba to accurately recognize samples from Class 1, although their accuracies on Class 2 are high.
These observations suggest that FreMGP can help Mamba improve the recognition accuracy on the minority class without degrading the recognition accuracy on the majority class.

\begin{figure}
    \centering
    \includegraphics[width=\columnwidth]{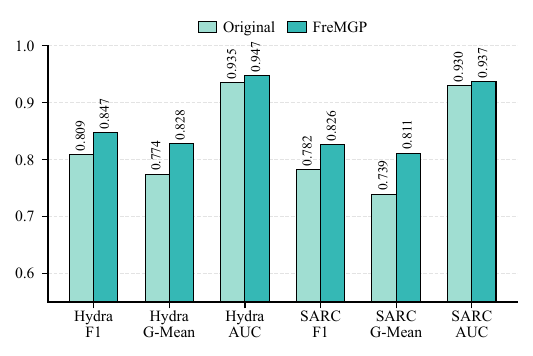}
    \caption{Performance comparison between the original imbalanced training data and FreMGP on additional classifiers.}
    \label{fig:additional_classifiers}
\end{figure}

\subsection{Performance Evaluation of FreMGP on State-of-the-Art (SOTA) Classifiers}

To further examine whether FreMGP can be applied to SOTA time-series classifiers, we evaluate its performance with Hydra~\cite{dempster2023hydra} and SARC~\cite{pmlr-v267-liu25by}. As shown in Fig.~\ref{fig:additional_classifiers}, FreMGP improves the performance of both classifiers over the original imbalanced training data across all the three evaluation metrics. For Hydra, FreMGP increases the F1-Score result from 0.809 to 0.847 and the G-Mean result from 0.774 to 0.828, while the AUC result shows a smaller improvement from 0.935 to 0.947. A similar trend can be observed for SARC, where FreMGP improves the F1-Score result from 0.782 to 0.826 and the G-Mean from 0.739 to 0.811, with a slight increase in AUC performance from 0.930 to 0.937. These results suggest that the samples generated by FreMGP can provide useful additional training information for different classifiers. 
The detailed results on each dataset are provided in Appendix D.4.

\subsection{Ablation Analysis}

\begin{table*}
\centering
\caption{Ablation study results under different classifiers on 14 datasets}
\label{tab:ablation}
\resizebox{\textwidth}{!}{
\begin{tabular}{llccccccc}
\toprule
Classifier & Metric
& w/o Gen. Ops.
& w/o Proto. Loss
& w/o Stage I
& w/o Stage II
& w/o Diversity
& w/o Similarity
& FreMGP \\
\midrule
\multirow{3}{*}{LSTM}
& F1-Score 
& 0.592 & 0.678 & 0.617 & 0.684 & \underline{0.695} & 0.678 & \textbf{0.718} \\
& G-Mean   
& 0.509 & 0.646 & 0.536 & 0.659 & \underline{0.671} & 0.647 & \textbf{0.705} \\
& AUC      
& 0.770 & 0.822 & 0.785 & 0.829 & \underline{0.834} & 0.822 & \textbf{0.846} \\
\midrule
\multirow{3}{*}{Transformer}
& F1-Score 
& 0.694 & 0.742 & 0.737 & \underline{0.754} & 0.750 & 0.748 & \textbf{0.762} \\
& G-Mean   
& 0.648 & 0.715 & 0.708 & \underline{0.733} & 0.729 & 0.731 & \textbf{0.745} \\
& AUC      
& 0.840 & 0.875 & 0.871 & 0.877 & 0.876 & \underline{0.878} & \textbf{0.897} \\
\midrule
\multirow{3}{*}{Mamba}
& F1-Score 
& 0.719 & 0.766 & 0.748 & 0.783 & \underline{0.792} & 0.783 & \textbf{0.804} \\
& G-Mean   
& 0.656 & 0.724 & 0.704 & 0.759 & \underline{0.773} & 0.758 & \textbf{0.788} \\
& AUC      
& 0.879 & 0.900 & 0.897 & 0.898 & \underline{0.905} & 0.904 & \textbf{0.917} \\
\bottomrule
\end{tabular}
}
\end{table*}

\begin{figure*}
\centering
\includegraphics[width=1\linewidth]{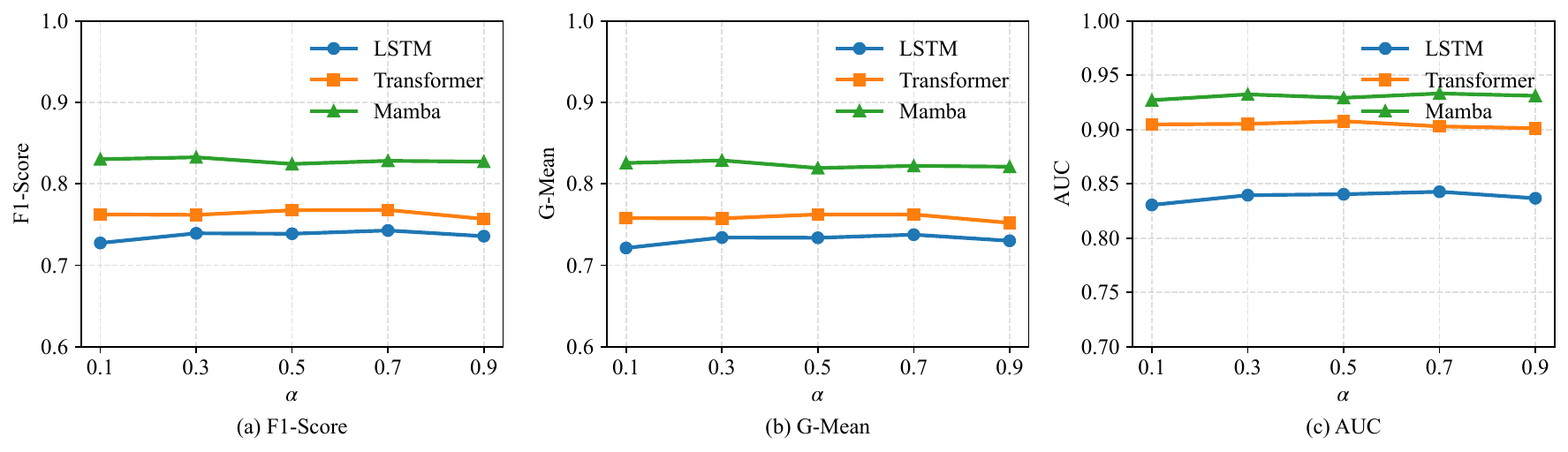}
\caption{Sensitivity analysis of the trade-off coefficient $\alpha$ in the Stage-II fitness function.}
\label{fig:alpha_sensitivity}
\end{figure*}

Table~\ref{tab:ablation} reports the ablation study results averaged over the 14 datasets under LSTM, Transformer, and Mamba classifiers. To examine the contribution of each component, we construct several variants of FreMGP. The variant ``w/o Gen. Ops.'' removes the proposed stage-specific genetic-operator design. The variant ``w/o Proto. Loss'' removes the prototype-guided loss from the representation learning module. The variants ``w/o Stage I'' and ``w/o Stage II'' remove the representation proximity search and the local distribution expansion search, respectively. The variants ``w/o Diversity'' and ``w/o Similarity'' remove the angular diversity term and the radial consistency term in the Stage-II fitness function, respectively. The detailed dataset-level results are provided in Appendix D.5.

As shown in Table~\ref{tab:ablation}, FreMGP achieves the best average results among all its variants across the three classifiers and all three evaluation metrics. When the proposed genetic operator design is removed, performance drops significantly, particularly under the LSTM classifier. This suggests that the stage-specific tree-level and group-level operators play an important role in effectively guiding the evolutionary search. Removing the prototype-guided loss also reduces the performance, suggesting that the prototype constraint contributes to learning a more useful representation space for fitness evaluation.

The results of ``w/o Stage I'' and ``w/o Stage II'' show that both search stages are useful. Without Stage I, the performance drops noticeably, especially in terms of F1-Score and G-Mean. Without Stage II, the performance is also lower than that of the full FreMGP, indicating that local distribution expansion further improves the quality of generated samples once representation proximity has been attained. 
In addition, removing either the angular diversity term or the radial consistency term leads to lower average performance compared with FreMGP. This suggests that considering both similarity and diversity in the fitness function of Stage II is beneficial for generating high-quality minority-class samples.

\subsection{Parameter Sensitivity Analysis}

To analyze the influence of the parameter $\alpha$ in the fitness function of Stage II, we vary $\alpha$ from 0.1 to 0.9 and conduct experiments under LSTM, Transformer, and Mamba classifiers. A larger $\alpha$ gives more importance to radial consistency, which encourages synthetic samples to stay close to the target minority samples. A smaller $\alpha$ gives more importance to angular diversity, which encourages synthetic samples to expand in different directions around the target samples.


Fig.~\ref{fig:alpha_sensitivity} reports the averaged results over six representative datasets to show the overall trend of FreMGP under different values of $\alpha$. Overall, FreMGP maintains stable performance when $\alpha$ varies from 0.1 to 0.9. This indicates that the parameter $\alpha$ is less sensitive. The detailed results on the six datasets are provided in Appendix D.6.

\section{Conclusions}

To generate high-quality and diverse minority-class time-series samples to mitigate the issue of class imbalance and enhance classifier performance for imbalanced time-series classification, we have proposed FreMGP, frequency-domain time-series samples for oversampling.
where a multi-tree GP structure has been carefully designed to represent frequency-domain time-series samples for oversampling. Furthermore, a frequency-domain contrastive representation learning module was designed, which provides a reliable representation space for evaluating synthetic samples. Based on this, a two-stage fitness evaluation strategy has been designed. The first stage drives generated samples toward the minority-class region in the learned representation space, while the second stage further encourages them to move in diverse directions by considering both similarity and diversity. This strategy effectively guides the evolutionary search toward samples that are close to minority-class samples in the learned representation space while ensuring sufficient diversity among the generated samples. 

Experiments on imbalanced time-series datasets have demonstrated that FreMGP improves the classification performance of various classifiers across different evaluation metrics, including F1-Score, G-Mean, and AUC. Nevertheless, FreMGP has a higher computational cost than traditional non-EC sampling methods, such as SMOTE, mainly due to the evaluation of individuals over multiple generations. In future work, we will explore more efficient evolutionary strategies to reduce the computational burden. 

\ifCLASSOPTIONcaptionsoff
  \newpage
\fi

\bibliographystyle{IEEEtran}
\bibliography{reference}

\end{document}